\pdfoutput=1

\documentclass[11pt]{article}
\PassOptionsToPackage{table}{xcolor}
\usepackage{microtype}
\usepackage{booktabs} 
\usepackage{xspace}
\usepackage{multirow}
\usepackage{graphicx}
\usepackage{subcaption}
\usepackage{booktabs}
\usepackage{wrapfig}
\usepackage{booktabs} 
\usepackage[dvipsnames]{xcolor}
\usepackage[table]{xcolor}
\usepackage{amssymb}
\usepackage{hyperref}
\usepackage{ulem}
\usepackage{amsmath}
\usepackage{amssymb}
\usepackage{mathtools}
\usepackage{amsthm}
\usepackage{xspace}
\usepackage{soul}
\usepackage[preprint]{acl}

\usepackage{times}
\usepackage{latexsym}

\usepackage[T1]{fontenc}

\usepackage[utf8]{inputenc}

\usepackage{microtype}

\usepackage{inconsolata}

\DeclareMathOperator{\softmax}{softmax}
\newcommand{\ie}{\textit{i.e.,}\xspace}
\newcommand{\eg}{\textit{e.g.,}\xspace}

\newcommand{\etc}{\textit{etc.}\xspace}

\newcommand{\modelname}{AdaInfer\xspace}

\newtheorem{observation}{\textbf{Observation}}

\title{Not All Layers of LLMs Are Necessary During Inference}

\author{
  Siqi Fan\textsuperscript{2},
  Xin Jiang\textsuperscript{1}, 
  Xiang Li\textsuperscript{1}, 
  Xuying Meng\textsuperscript{3},
  Peng Han\textsuperscript{2}, 
  Shuo Shang\textsuperscript{2*}, \\
  \textbf{
  Aixin Sun\textsuperscript{4}, 
  Yequan Wang\textsuperscript{1*},
  Zhongyuan Wang\textsuperscript{1}}\\\\
  $^{1}$Beijing Academy of Artificial Intelligence, Beijing, China\\
  $^{2}$University of Electronic Science and Technology of China, Chengdu, China\\
  $^{3}$Institute of Computing Technology, Chinese Academy of Sciences, Beijing, China\\
  $^{4}$College of Computing and Data Science, Nanyang Technological University, Singapore
}

\begin{document}
\maketitle

\renewcommand{\thefootnote}{\fnsymbol{footnote}}
\footnotetext[1]{Corresponding authors.}
\renewcommand{\thefootnote}{\arabic{footnote}}

\begin{abstract}
Due to the large number of parameters, the inference phase of Large Language Models (LLMs) is resource-intensive. However, not all requests posed to LLMs are equally difficult to handle. Through analysis, we show that for some tasks, LLMs can achieve results comparable to the final output at some intermediate layers. That is, \textit{not all layers of LLMs are necessary during inference}. If we can predict at which layer the inferred results match the final results (produced by evaluating all layers), we could significantly reduce the inference cost. To this end, we propose a simple yet effective algorithm named \textbf{\modelname} to adaptively terminate the inference process for an input instance. \modelname relies on easily obtainable statistical features and classic classifiers like SVM. Experiments on well-known LLMs like the Llama2 series and OPT, show that \modelname can achieve an average of 17.8\% pruning ratio, and up to 43\% on sentiment tasks, with nearly no performance drop (<1\%). Because \modelname does not alter LLM parameters, the LLMs incorporated with \modelname maintain generalizability across tasks.
\end{abstract}

 \section{Introduction}
\label{sec:intro}

LLMs have demonstrated impressive performance on various downstream tasks using evaluation protocols such as zero-shot, few-shot, and fine-tuning \cite{todd2023function,chan2022data,kossen2023context,wang2023label, DBLP:conf/emnlp/WangZSM22}. Example applications include text generation, question answering, and sentiment analysis. Notably, the in-context learning ability allows LLMs to adapt to various different tasks using input-output examples without parameter updates \cite{kossen2023context,todd2023function}. However, the inference phases of LLMs are typically very expensive due to their large number of parameters \cite{pope2023efficiently,liu2023deja}. Specifically, the inference time complexity for typical large models with a Transformer structure is $LSd(d+S)$ per single inference, where $L$, $S$, and $d$ represent the number of layers, sequence length, and hidden size, respectively.

Existing solutions to achieve more efficient inference in LLMs include model pruning \cite{ma2023llm,kim2024shortened} and sparse models \cite{lecun1989optimal, liu2023deja}. Both solutions alter LLM parameters and may risk compromising generalization ability. Additionally, different LLM designs pose compatibility challenges with other acceleration methods. Hence, an ideal efficient LLM inference should use fewer computational resources while maintaining generalization and in-context learning abilities \cite{liu2023deja}.

If we draw an analogy between LLM inference and the human thinking process~\cite{salthouse1996processing, deary2001reaction}, where simple questions can be answered quickly and complex questions require more time for reasoning, we may expect LLMs not to use the same inference power to handle all tasks. \citet{teerapittayanon2016branchynet, huang2017multi} show that "easy" tasks activate at shallower layers while "hard" ones activate at deeper layers. For LLM training, a growth strategy~\cite{DBLP:journals/corr/abs-2309-03852} adds parameters in stages to reduce the overall training cost, \ie not all training instances use the same set of parameters. Hence, we consider that adjusting the parameters during inference based on the difficulty level of a task may be an effective way for efficient inference.

To this end, we conduct a statistical analysis to examine the correlation between the results obtained in intermediate layers and those in the final layers across various tasks. We made two observations: (i) not all layers of LLMs are necessary during inference, \ie early stopping works, and (ii) simpler tasks require fewer layers, while more complex tasks require more layers of inference. The key to achieving efficient LLM inference then becomes \textit{when to stop the inference process adaptively based on the input instance}. Interestingly, exploring adaptive inference may bridge LLMs with the brain's information processing \cite{hubel1962receptive, murata2000selectivity}, aiding in the analysis of activated network modules during sample processing \cite{han2021dynamic} and identifying crucial input components that affect the final prediction.

In this paper, we present \modelname, a simple yet effective algorithm for instance-aware adaptive inference. The core of \modelname lies in data-driven decision-making. We begin by performing a statistical analysis on each block feature of LLMs, such as logits, hidden states, mlp, and attention activation values. Consequently, we choose logits to construct features and employ classical statistical classifiers, SVM and CRF, to predict the optimal layer at which to stop the inference. Experiments on well-known LLMs (\ie Llama2 series and OPT) show that \modelname can achieve an average of 17.8\% pruning ratio, and up to 43\% on sentiment tasks, with nearly no performance drop (<1\%). The cost of collecting the small set of statistical features and running \modelname is negligible compared to the cost of LLM inference.

\modelname is an early stop strategy that optimizes efficiency without altering the model's parameters. Therefore, \modelname does not affect the model's generalization and in-context learning abilities. Furthermore, being orthogonal to other model acceleration techniques, \modelname offers the potential for further enhancing inference efficiency.

\section{Related Work}
\label{sec:related}

Existing solutions for achieving adaptive inference involve dynamic neural networks \cite{han2021dynamic, huang2017multi, bolukbasi2017adaptive}. These solutions can be broadly classified into two groups: dynamic depth (number of network layers) and dynamic width (number of channels, parallel subnetworks, \etc).

\paragraph{Dynamic Depth} involves two methods: \textit{Early Exit (EE)} and \textit{Skip Layer}. \textit{EE} first appeared in CNN/DNN networks for visual tasks \cite{bolukbasi2017adaptive, huang2017multi, teerapittayanon2016branchynet}. Subsequently, it was utilized to accelerate the inference of encoder-only architectures in BERT \cite{li2020cascadebert,liu2020fastbert,li2021accelerating,kong2022accelerating}. Recently, \citet{schuster2022confident, varshney2023accelerating} discussed confidence-based \textit{EE} for LM adaptive inference. Our proposed \modelname closely aligns with the \textit{EE} concept. We apply \textit{EE} to mainstream decoder-only LLMs, which adhere to the scaling law but suffer from high inference costs due to their large parameter count.

\textit{Skip Layer} dynamically omits the execution of middle layers (or modules) for an input token, facilitated by a gate function \cite{wang2018skipnet}, a binary router \cite{zeng2023learning,raposo2024mixtureofdepths}, or layer pruning \cite{kim2024shortened,yang2024laco,song2024sleb,men2024shortgpt, ma2023llm}. The main difference between our method and theirs is that we achieve instance-wise inference (\ie dynamic pruning ratio tailored to specific tasks) without altering the model parameters, which is crucial for current LLMs. To the best of our knowledge, this is the first attempt to discover that each block's logits are crucial elements for \textit{EE} classifiers in LLMs, and we incorporate this insight as a fundamental design choice in \modelname.

\paragraph{Dynamic Width} controls the number of neurons in the network width for efficient inference. This includes methods such as reducing the number of CNN channels \cite{hua2019channel, hoefler2021sparsity} and establishing multiple parallel structures for "experts" in Mixture of Experts (MoE) \cite{fedus2022switch, zhou2022mixture, artetxe2021efficient}, dynamically weighting and predicting the output results. Recently, \citet{ma2023llm, addanki2023one, xia2023sheared} have slimmed the network width by pruning attention heads and the output neurons in Query, Key, or Value. Other model acceleration methods, such as quantization \cite{xiao2023smoothquant, xing2023bipft} and sparsity \cite{liu2023deja,frantar2023sparsegpt}, are orthogonal areas and usually excel in different settings.

\section{Efficiency Analysis of LLM Inference}
\label{sec:analysis}
Before presenting the statistical observations and insights on LLM inference, we first briefly review LLM's critical components.

\subsection{Preliminary: LLM Building Blocks}
\label{ssec:llmPrelim}

Modern LLMs, rooted in the Transformer architecture \cite{DBLP:conf/nips/VaswaniSPUJGKP17}, can be trained with various unsupervised training objectives. In this paper, we focus on mainstream LLMs like GPT and the Llama series. These models are built with a decoder-only structure and are pre-trained with a full language modeling objective, computing loss on all tokens.  Their key components can be broken down into the following blocks: \textit{Tokenizer and Embedding Layer}, \textit{Decoder Block}, and \textit{Classification Layer}. The tokenization and embedding layer converts input text into numerical vectors, enabling effective processing and analysis of textual data. The decoder block processes numerical vectors through self-attention and feedforward neural networks, allowing the model to focus on, or attend to, the most relevant parts of the input. Lastly, the classification layer, or the LM head layer, maps decoder logits into a vocabulary-wide probability distribution to facilitate word prediction. These blocks facilitate LLMs in efficiently handling NLP downstream tasks, with a primary emphasis on the decoder block. 

During inference, each input instance passes through the decoder block, layer by layer, until the last layer. Hence, the inference complexity is linearly related to the number of decoder layers $L$ in the decoder block. The decoder block of earlier models typically comprised 6 layers, whereas current open-source models have many more. For example, Llama2-7B has 32 layers and Llama2-13B features 40 decoder layers \cite{touvron2023llama}.

\begin{figure}
    \centering
    \includegraphics[width=0.75\linewidth]{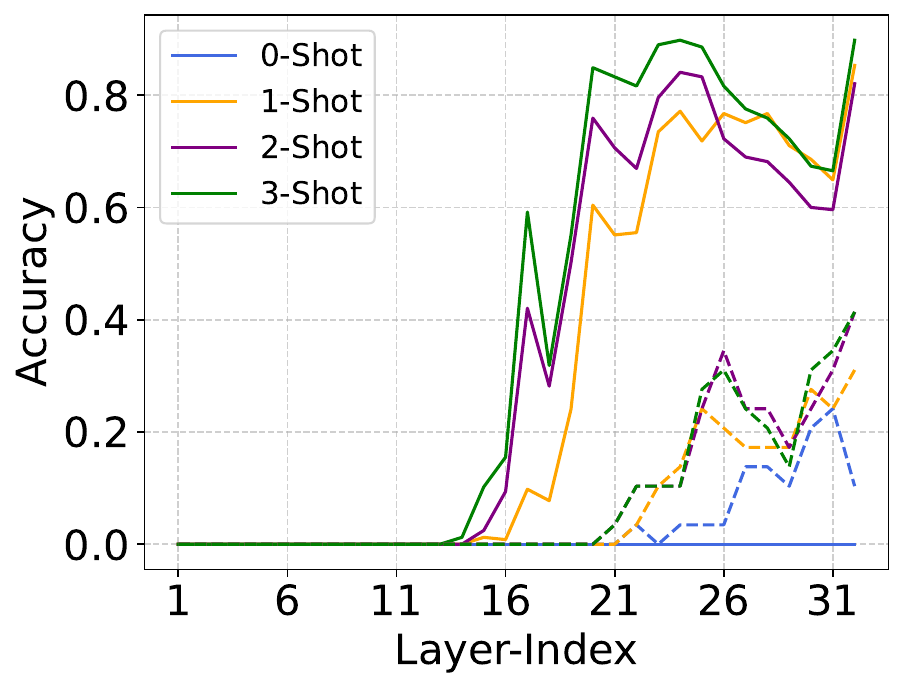}
    \caption{Accuracies obtained by inference at each decoder layer with the Llama2-7B model (32 layers). The \textit{solid line} represents the sentiment analysis task, and the \textit{dashed line} represents the MMLU task.}
    \label{fig:simple_hard}
\end{figure}

\subsection{Not all Layers are Necessary}
\label{not all layers are necessary}

To explore the possibility of skipping some intermediate layers during inference, we conduct experiments on two tasks: sentiment analysis \cite{socher-etal-2013-recursive} and MMLU \cite{hendryckstest2021}. We examine the accuracies obtained by stopping inference at each decoding layer and compare them with the final results, \ie without stopping inference. The experiments were conducted on both Llama2-7B (32 layers) and Llama2-13B (40 layers), and the same observations hold. 


\begin{observation}\label{obs:eeworks}
Not all layers of LLMs are necessary during inference: Early Stopping works.
\end{observation}
Using the SST-2 dataset \cite{socher-etal-2013-recursive}, we conduct sentiment classification experiments on the Llama2-13B (40 layers) model. We perform inference at each layer with a batch size of 1 and record the results. On average, an early exit at layer 21 (with a variance of 5.1) achieves comparable accuracy to the final layer output. Interestingly, simpler inputs like `I like Camera A' activate only 18 layers, while more complex inputs like `Camera A is better than Camera B in picture quality' activate about 24 layers. Early stop works on the Llama2-7B model as well. 

\begin{observation}\label{obs:tasks}
Simpler tasks require fewer layers for inference, while complex tasks go deeper.
\end{observation} 
Figure~\ref{fig:simple_hard} plots the accuracies by stopping inference at different decoding layers on a Llama2-7B. For the task of sentiment analysis, the accuracy matches that of the final layer by the 24th layer, represented by solid lines in the figure. For MMLU, a complex task, accuracy tends to improve with deeper layers. A similar trend holds across all four tested settings, from 0-shot to 3-shot learning. 

\paragraph{Insight.}
Both observations are intuitive and, in fact, not new. Similar findings have been made in visual tasks with convolutional neural networks \cite{teerapittayanon2016branchynet, huang2017multi} and sentence classification with BERT \cite{liu2020fastbert}. We extend these observations to decoder-only LLM inferences. 

Based on the two observations, we understand that (i) early stopping works, allowing us to reduce inference costs by stopping at certain decoding layers without compromising model accuracy, and (ii) the number of optimal decoding layers for inference is instance-dependent. The number of optimal decoding layers varies across tasks and even across instances of the same task. Recall the two example sentences for sentiment analysis discussed in Observatoin~\ref{obs:eeworks}. This means that the layer at which inference stops must be dynamically determined (or predicted) for each input instance.

\begin{figure}
  \centering
  \begin{subfigure}[b]{\linewidth}
    \centering
    \includegraphics[width=\linewidth]{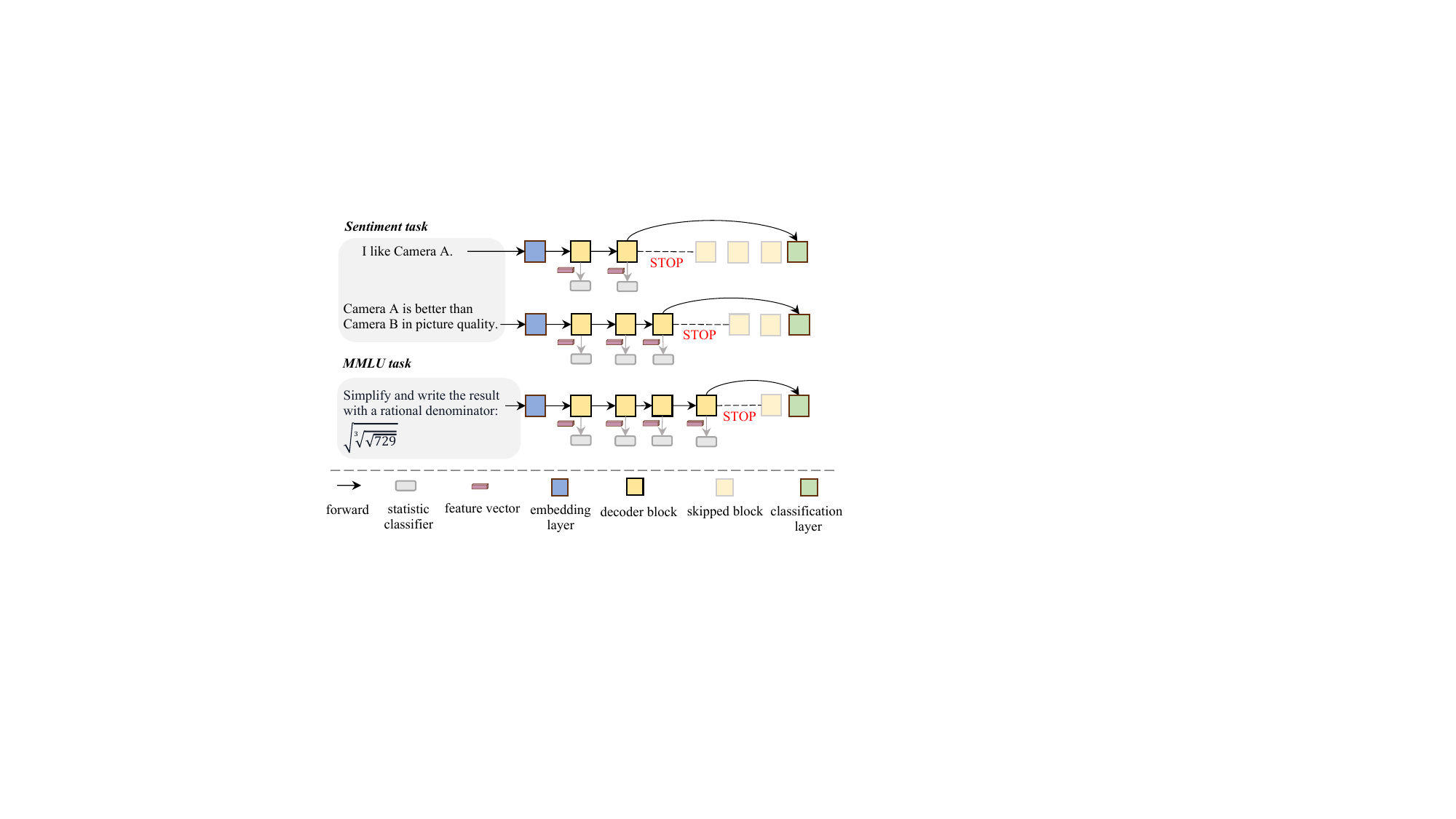}
    \caption{\modelname processes three input instances for two tasks, with inference stopping at different decoding layers.}
    \label{sfig:process}
  \end{subfigure}
  
  \begin{subfigure}[b]{\linewidth}
    \centering
    \includegraphics[width=\linewidth]{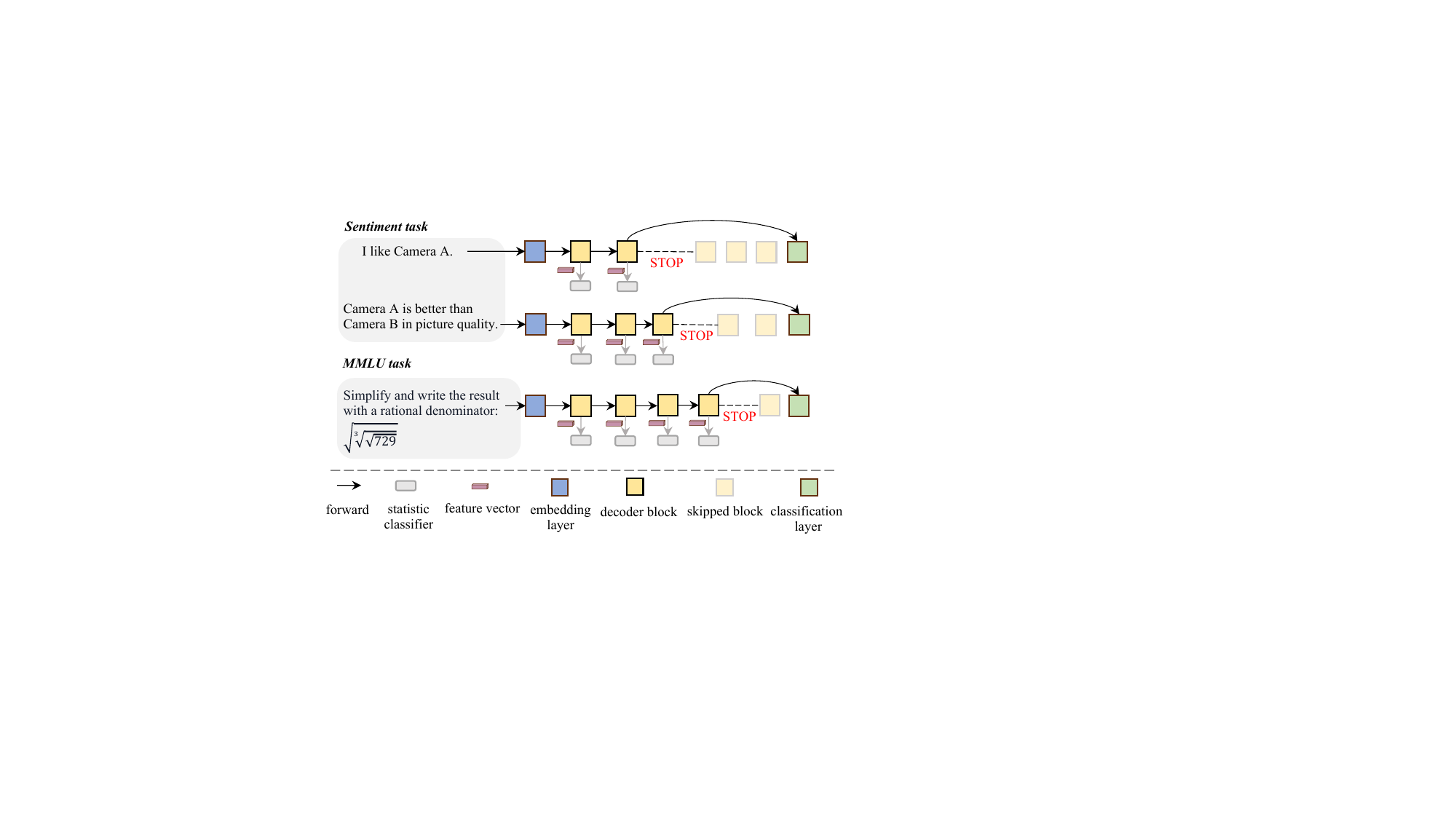}
    \caption{Effectiveness in reducing computational costs with early stopping during inference.}
    \label{sfig:statisticEarlyExit}
  \end{subfigure}
  \caption{An illustration of \modelname's processing and computational savings.}
  \label{fig:overview}
\end{figure}

\section{\modelname: Adaptive Inferences}
\label{method}
Modifying LLM parameters may require additional training and pose a potential risk of compromising the model's generalization capabilities \cite{gu2024model}. In designing \modelname, we embrace a cost-effective approach that preserves the model's innate abilities without altering its parameters. The main idea is to capture signals at each decoding layer and make predictions on whether to stop the inference at the current layer. 

The workflow of \modelname is depicted in Figure~\ref{sfig:process} with three example input instances. At each decoding layer, a \textit{Feature Selection} component crafts a feature vector for the current input instance. A binary \textit{Classifier} then predicts whether to stop the inference, \ie bypass subsequent decoder layers.

\subsection{Feature Selection}
\label{ssec:features}

LLMs capture coarse-grained features in their initial layers and develop more detailed, fine-grained representations in deeper layers. This process is facilitated by the repeated application of multi-head attention mechanisms and the use of residual connections. However, there is a lack of features to demonstrate at which stage the representation is sufficient for the current task. Furthermore, these features need to be inherently universal to ensure compatibility across various LLMs.

\begin{figure}
  \begin{subfigure}{0.48\linewidth}
    \centering
    \includegraphics[width=\linewidth]{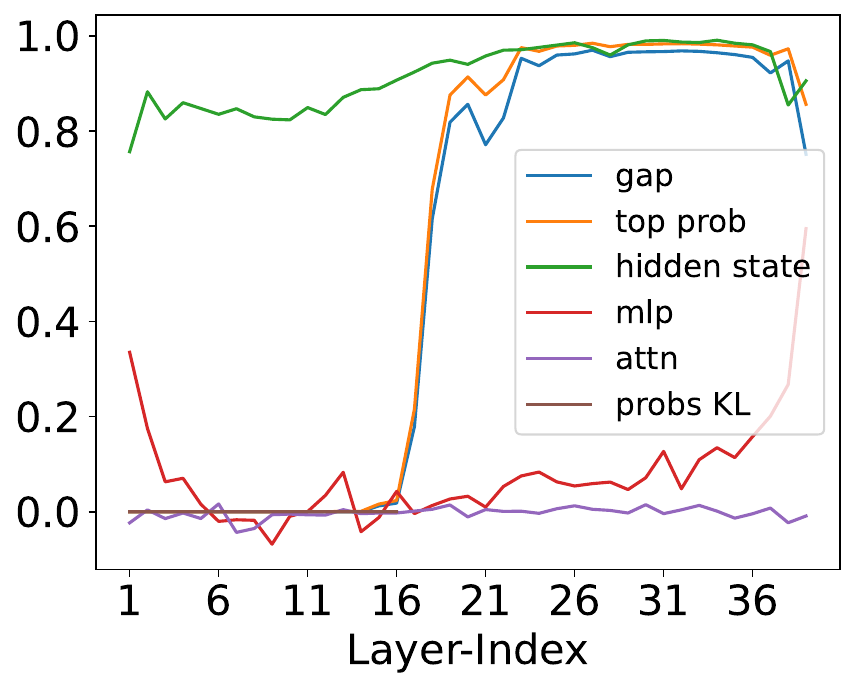}
    \caption{Sentiment Task}
    \label{sfig:sentiment}
  \end{subfigure}
  \begin{subfigure}{0.48\linewidth}
    \centering
    \includegraphics[width=\linewidth]{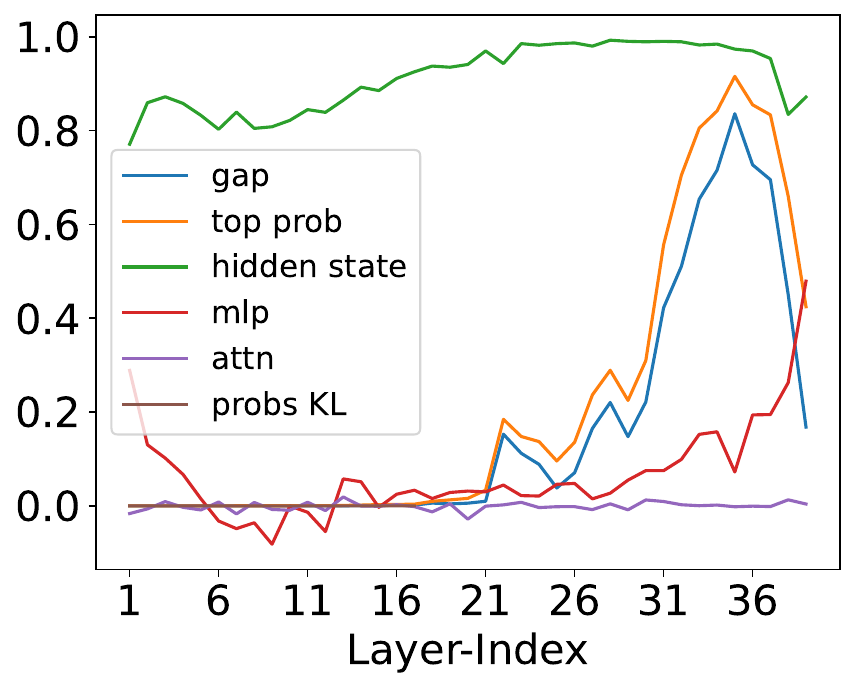}
    \caption{MMLU Task}
    \label{sfig:MMLU}
  \end{subfigure}
  \caption{Changes of feature values along the 40 decoding layers in Llama2-13B model.}
  \label{fig:featureAnalysis}
\end{figure}


As a part of feature engineering, we conduct a visual analysis of diverse features from each decoding layer (or decoding block illustrated in Figure~\ref{sfig:process}) of LLMs. Our examination focused specifically on:
\begin{itemize}
    \item \textbf{Gap} measures the current block's prediction confidence for the next token, defined as $P(\text{top token})-P(\text{second token})$, where $P$ represents the probability distribution generated by the current block.
    \item \textbf{Top Prob}, $P(\text{top token})$, is the probability estimation of the most likely next token by the current block. 
    \item \textbf{Cosine Similarities} between the current and the previous blocks, calculated on attention activation value (attn), multi-layer perceptron outputs (mlp), and hidden states, respectively.
\end{itemize}

Again, we use the sentiment and MMLU tasks on the Llama2-13B (40 layers) model for feature analysis, shown in Figure~\ref{fig:featureAnalysis}. Observe the following trends: (1)  across tasks, both ``gap'' and ``top prob'' gradually increase alone the inference phase, stabilizing in the deeper layers. (2) The activation of ``gap'' and ``top prob'' varies across layers for different tasks. These phenomenons are also evident in the Llama2-7B, OPT-13B \cite{zhang2022opt}, and GPT-J \cite{gpt-j} (See Appendix \ref{more llm observation}). The feature analysis suggests that \textbf{``gap'' and ``top prob'' can serve as universal features} for the inference-stopping signal. Notably, these two values remain consistent across two diverse tasks, indicating a versatile discriminating power applicable to various tasks. Factor studies in subsequent experiments also show that other features (\eg cosine similarities) exhibit subtle differences across layers.

\subsection{Classifier}
\label{ssec:classifier}

The classifier determines if the signal is compelling enough to warrant an early termination of the process. There are many choices for classifiers, ranging from rule-based classifiers~\cite{huang2017multi, yang2020resolution, DBLP:conf/emnlp/WangZSM22} to gating functions~\cite{lin2017runtime, bejnordi2019batch}. In our context, classical statistical classification methods are a good option due to their efficiency and their ability to handle simple input features (\ie ``gap'' and ``top prob'') for a binary classification task.

Given one instance, we obtain the feature vector $x_d$ using the feature selection module. If the current layer's output $\hat{y}$ provides the correct answer $y$, the associated label $y_c$ is a positive example; otherwise, it is a negative example. For LLMs trained to predict the next token, if the next token $\hat{y}$ predicted based on an intermediate decoding layer's output is the same as the token $y$ predicted by the last decoding layer's output, then the layer's label $y_c=1$.

\begin{equation}
    y_c = \begin{cases} 
    1 & \text{if } \hat{y} = y, \\
    0 & \text{otherwise}.
    \end{cases}
\end{equation}
Thus, for an $L$-layer LLM, each input instance $x$ yields $L$ pairs of $\langle x^d, y_c\rangle$. The details of creating training data for the classifier are provided in Appendix~\ref{sec:createTrainingData}. In our implementation, we consider two types of classifiers: Support Vector Machines (SVM) \cite{hearst1998support} and Conditional Random Fields (CRF) \cite{lafferty2001conditional}. SVM does not rely on the context of sequences, while CRF incorporates sequence modeling along the decoding layers.

\section{Experiments}
\label{sec:experiment}
\definecolor{myblue}{HTML}{EDC5AB}

We now conduct experiments with \modelname on well-known LLMs across various tasks. Specifically, we evaluate the zero/few-shot learning capabilities, with two primary types of tasks.

\paragraph{Question Answering Tasks.} (1) MMLU \cite{hendryckstest2021} encompasses 57 tasks across humanities, social sciences, STEM, and more, requiring world knowledge and problem-solving capabilities. (2) CommonsenseQA \cite{talmor-etal-2019-commonsenseqa} tests for commonsense knowledge through multiple-choice questions. (3) SQuAD \cite{2016arXiv160605250R} serves as a reading comprehension benchmark, with questions based on Wikipedia articles and answers either segments of passage or marked as unanswerable.

\paragraph{Text Classification Tasks.} (1) SST-2 \cite{socher-etal-2013-recursive} involves sentiment analysis of movie reviews with binary ``positive'' or ``negative'' labels. (2) AG News \cite{zhang2015character} classifies news headlines and article sentences into Business, Science/Technology, Sports, and World categories.

\subsection{Experiment Settings} 
\label{ssec:expSetting}

\begin{table}
    \centering
    \caption{LLMs used in experiments with \modelname.}
    \begin{tabular}{lccc}
    \toprule 
    Model &Params & Tokens & Layer Num.  \\
    \midrule
    Meta/OPT  & 13B & 0.18T & 40 \\
    Meta/Llama 2 & 7B & 2T & 32  \\
    Meta/Llama 2 & 13B & 2T & 40  \\
    Meta/Llama 2 & 70B & 2T & 80  \\
    \bottomrule
    \end{tabular}
    \label{tab:model} 
\end{table}

\definecolor{lightblue}{RGB}{173, 216, 230}
\begin{table*}[tb]
    \centering
    \caption{Performance and computational efficiency in multi-tasks. Accuracy (\%) is denoted by ‘Acc’. Results of few-shot learning with sample sizes of \{5, 10, 15, 20\} are reported in average values. ShortGPT$_p$ follows the orignal paper's setting;  ShortGPT$_5$ and ShortGPT$_3$ are to skip the last 5 and 3 decoding layers, respectively.}
    \setlength{\tabcolsep}{1mm}
    \resizebox{1\textwidth}{!}{
        \begin{tabular}{l|c|ccc|ccc|ccc|ccc|ccc}
            \toprule
            \multirow{2}{*}{Tasks} &\multirow{2}{*}{P. Ratio(\%)}&  \multicolumn{3}{@{}c}{{\bf MMLU}} &  \multicolumn{3}{@{}c}{{\bf CommonsenseQA}} & \multicolumn{3}{@{}c}{{\bf SQuAD}}  &  \multicolumn{3}{@{}c}{{\bf Sentiment}}&\multicolumn{3}{@{}c}{{\bf AG News}}\\
            && Acc & \#Avg. L & Var & Acc & \#Avg. L & Var & Acc & \#Avg. L & Var& Acc&\#Avg. L & Var& Acc & \#Avg. L & Var\\
            \cmidrule (r){1-1}\cmidrule (lr){2-2} \cmidrule (lr){3-5} \cmidrule (lr){6-8} \cmidrule (lr){9-11} \cmidrule (lr){12-14}\cmidrule (lr){15-17}

            \multicolumn{17}{@{}l}{{\bf \textit{Llama 7B (32 Layers)}}}  \\
            Dense   &--&43.05 &32&-- &53.50&32&-- &48.08&32&-- &95.20&32&-- &79.65&32&-- \\
			ShortGPT$_p$ &28.13&21.52&23&--&33.52&23&--&10.60&23&--&93.48&23&--&56.90&23&-- \\ 
			ShortGPT$_5$ &15.63&29.95&27&--&41.90&27&--&12.97&27&--&90.40& 27&--&53.25&27&--\\
            ShortGPT$_3$ &9.38& 37.39&29&--&53.22&29&--&14.32&29&--&94.17&29&--&71.28&29&--\\
            \rowcolor{lightblue!30} \textbf{\modelname}  &9.66 $\rightarrow$ 35.71&43.73 &28.91&4.97&53.00&27.90&5.93&45.82&26.77&11.88&95.30&20.57&5.10&79.72&29.20&2.70  \\
            \midrule
            \midrule
            \multicolumn{17}{@{}l}{{\bf \textit{Llama 13B (40 Layers)}}}  \\ 
            Dense   &--&53.31&40&--&64.92&40&--&52.90&40&--&95.90&40&--&77.53&40&-- 
            \\
            ShortGPT$_p$ &25&45.12&30&--&65.00&30&-- &13.32&30&--&84.38&30&--&55.90&30&--\\
           ShortGPT$_5$ &12.50&46.64&35&--&64.45&35&--&16.35 &35&--&89.80&35&--& 70.17&35&-- \\
           ShortGPT$_3$ &7.50&47.22&37&--&64.47&37&--&17.25&37&--&95.90&37&--&75.47&37&--\\
           \rowcolor{lightblue!30}  \textbf{\modelname} &9.13 $\rightarrow$ 43.33&52.44&36.35&8.15&62.48&34.60&10.20&48.35&31.18&31.75&92.65&22.67&8.10&76.43&34.02&24.18  \\
            \midrule
            \midrule
            \multicolumn{17}{@{}l}{{\bf \textit{OPT 13B (40 Layers)}}}  \\ 
            Dense   &--&23.60&40&--&21.45&40&--&26.12&40&--&92.58&40&--&72.83&40&-- 
            \\
            ShortGPT$_p$ &25&10.17 &30&--  &11.50&30&--&0.65&30&--&14.72&30&--&2.27&30&--\\
           ShortGPT$_5$ &12.50& 22.92&35&--&19.12&35&--&22.12&35&--&86.33&35&--&49.42&35&--\\
           ShortGPT$_3$ & 7.50& 23.05 &37&--& 19.68 &37&--&24.65&37&--&91.35&37&--&66.62&37&--\\
           \rowcolor{lightblue!30}  \textbf{\modelname} &9.75 $\rightarrow$ 22.63&22.59&32.37&7.92&21.62&33.33&12.12&25.95&34.20&13.50& 92.97&30.95&5.77&72.83&39.00&0.00 \\

            \bottomrule
            \end{tabular}
    }
    \label{tab:main_results}
\end{table*}

\paragraph{Large Language Models.} For \modelname's backbone, we choose widely recognized LLMs, \ie  OPT \cite{zhang2022opt} and the Llama 2 series \cite{touvron2023llama}, detailed in Table~\ref{tab:model}. These models vary in terms of the number of parameters, ranging from 7B to 70B, and the number of layers, ranging from 32 layers to 80 layers.  

\paragraph{In-Context Learning Setting.} We evaluate \modelname under zero-shot and few-shot scenarios, using sample sizes of 5, 10, 15, and 20. For zero-shot, the input is the test set's $x_q$. For few-shot, training set examples are added to $x_q$. For in-context learning prompts, we use a default template: $\mathrm{Q}:\{x_{k}\} \backslash \mathrm{n} \mathrm{A}:\{y_{k}\} \backslash \mathrm{n} \backslash \mathrm{n}$, concatenating random $x_k$ and $y_k$ samples from task-specific training sets.

\paragraph{Metrics.} We report the top-1 accuracy score on the test set following function vectors \cite{todd2023function} (HELM implementation)\footnote{\url{https://huggingface.co/blog/open-llm-leaderboard-mmlu}}. For computational efficiency, we follow previous work \cite{ma2023llm,schuster2022confident,elbayad2019depth} and report the pruning ratio (\textit{P. Ratio}) and the average number of activated layers (\textit{\#Avg. L}) for each task, along with their variance (\textit{Var}).
These metrics directly measure complexity reduction, avoiding conflation with implementation or infrastructure-specific details \cite{dehghani2021efficiency}.
For reference, we also translate them into FLOPs reduction ratios, in the Appendix \ref{all llms results}. Considering the conditional checks and classifier computation involved in \modelname, we also compare the actual speed of \modelname in real-world scenarios with Dense implementation, reporting wall-clock time \cite{dehghani2021efficiency}. 

\paragraph{Baseline Method: ShortGPT.}
We compare \modelname with the structured pruning method ShortGPT \cite{men2024shortgpt}, which prunes redundant layers in LLMs based on similarity scores. For the OPT model, we calculate redundant layers as outlined in the paper. For the LLama model, we use the same layers reported. Note that these model pruning methods apply a static pruning ratio across all tasks, whereas our \modelname adaptively performs model pruning based on input.

\subsection{Main Results} 
\label{main results}
The main results of \modelname are presented in Table~\ref{tab:main_results}. Conducted in few-shot settings, these experiments show the Top-1 accuracy, pruning ratios, average active layers for each task, and their variance. From the perspective of performance and computational efficiency, we draw the following experimental conclusions.

\paragraph{\modelname has minimum impact on performance (<1\%).} 
Table~\ref{tab:main_results} shows that the Top-1 accuracy of \modelname remains within a very narrow margin of less than 1\% for all tasks compared to dense models, \ie without early exit. In contrast, ShortGPT, following the paper's setting and denoted as ShortGPT$_p$, experiences a significant performance drop for almost all tasks \footnote{We noted a decline in the performance of the reproduced ShortGPT on the SQuAD dataset when the prompts increased to {10, 15, 20} shots.}. Since AdaInfer adaptively skips decoding layers, the number of layers skipped varies for different instances and across different tasks. For a fair comparison, we have also evaluated ShortGPT$_5$ and ShortGPT$_3$, which skip the last 5 and 3 decoding layers, respectively. The numbers of skipped layers are chosen to match the overall range of layers skipped by \modelname. This allows for a more comprehensive comparison with methods that use a fixed pruning ratio \cite{yang2024laco, ma2023llm,men2024shortgpt}. The results in Table~\ref{tab:main_results} demonstrate that AdaInfer surpasses both settings.

In short, \modelname achieves adaptive inference while maintaining LLM capabilities and in-context learning abilities without modifying model parameters. This finding is promising, especially in light of our Observation~\ref{obs:eeworks}, where we demonstrate the feasibility of implementing early exit strategies while preserving performance. As shown in Table~\ref{tab:main_results}, \modelname even surpasses the last layer accuracy for certain tasks. This suggests that deep layers may over-represent certain instances, potentially impeding performance during LLM inference.

\paragraph{Pruning ratio ranges from 9\% to 43\%, average 17.8\%.} We report the average and variance of the activated layers for each task and compute the pruning ratios in Table~\ref{tab:main_results}. The pruning ratios vary from task to task, ranging from $9\%$ to $43\%$, a clear indication of \modelname assesseing different early exit layer configurations for different task inputs. More layers are skipped for simple tasks like sentiment analysis task, where a $43\%$ reduction in computational cost can be achieved on Llama2-13B. For more complex question answering tasks, the savings range from $9\%$ to $20\%$. 

\paragraph{Wall-clock time.}
\begin{table}
    \centering
    \caption{Wall-clock time (s) and actual speedup for 358 test samples from MMLU and 245 test samples from Sentiment Tasks.}
    \resizebox{0.5\textwidth}{!}{
    \begin{tabular}{lcccccc}
    \toprule
    \multirow{2}{*}{Task} & \multicolumn{3}{c}{Llama2 7B (FP32)} & \multicolumn{3}{c}{{Llama2 13B (FP16)}} \\
        \cmidrule(lr){2-4} \cmidrule(lr){5-7}
        & Dense & \textbf{Ours} &Speed up& Dense & \textbf{Ours} &Speed up\\
    \midrule
    MMLU & 796.53 & 781.31 &1.02x& 339.19 & 320.46&1.05x \\
    Sentiment & 41.18 & 39.69  &1.04x& 28.18 & 21.76 &1.30x\\
    \bottomrule
    \end{tabular}}
    \label{tab: wall clock}
\end{table}

Next, we study the end-to-end runtime of \modelname. Table~\ref{tab: wall clock} compares the runtime of \modelname with a dense implementation on MMLU and Sentiment tasks (5-shot, batch size set to 1), using $6 \times V100$ (32GB).
We observed a 1.03x speed up on MMLU and 1.17x speed up on Sentiment when applying \modelname. That is, despite \modelname converting hidden states to logits at each block through the LM head layer, it only utilizes the last token's hidden state, which is independent of the input sequence length. Consequently, this computation cost is minimal (0.03\% of the total FLOPs for transformer inference). Further computational details on this process can be found in Appendix \ref{tf cost}. Meanwhile, statistical classifiers like SVM have much lower computational costs compared to LLM inference, as detailed in Appendix \ref{svm cost}, highlighting the computational efficiency potential of \modelname.

\begin{table}
    \centering
    \caption{Comparative analysis of GAP and CRF on performance and computational efficiency.}
    \resizebox{\linewidth}{!}{
    \begin{tabular}{c|c|cc|cc}
        \toprule
        \multirow{2}{*}{Task} & \multirow{2}{*}{Setting} & \multicolumn{2}{c|}{\modelname \textit{w.} Rule} & \multicolumn{2}{c}{\modelname \textit{w.} CRF} \\
        \cmidrule{3-6} 
        & & Acc$\uparrow$ & FLOPs$\downarrow$ & Acc$\uparrow$ & FLOPs$\downarrow$ \\
        \midrule
        \multirow{2}{*}{MMLU} & Zero-shot & 5.35 & 90.84 & 4.77 & 97.40 \\
        & Few-shot & 47.09 & 84.10 & 52.72 & 97.15 \\
        \midrule
        \multirow{2}{*}{CommonsenseQA} & Zero-shot & 1.10 & 92.78 & 1.40 & 97.28 \\
        & Few-shot & 55.33 & 79.57 & 65.72 & 96.40 \\
        \midrule
        \multirow{2}{*}{SQuAD} & Zero-shot & 24.60 & 73.17 & 23.10 & 93.03 \\
        & Few-shot & 43.43 & 71.19 & 51.75 & 89.94 \\
        \midrule
        \multirow{2}{*}{Sentiment} & Zero-shot & 0.00 & 88.25 & 0.00 & 97.27 \\
        & Few-shot & 91.45 & 51.25 & 95.60 & 73.07 \\
        \midrule
        \multirow{2}{*}{AG News} & Zero-shot & 0.10 & 77.82 & 0.10 & 94.04 \\
        & Few-shot & 69.17 & 70.65 & 76.77 & 93.08 \\

        \bottomrule
    \end{tabular}
    }
    \label{tab: gap and crf}
\end{table}

\subsection{Evaluation on Alternative Exit Strategies}
In Table~\ref{tab:main_results}, we employ SVM as the classifier for \modelname. To explore the impact of alternative exit strategies, Table~\ref{tab: gap and crf} reports \modelname implemented with a GAP threshold set at 0.8 (stopping inference when the current block's GAP feature exceeds 0.8) and \modelname with CRF as the classifier. The results show that both GAP and CRF can reduce computational costs by 3\% to 50\% while maintaining comparable LLM performance. Notably, in the zero-shot setting, GAP outperforms CRF, suggesting a relatively weak dependency between block features.

\subsection{Evaluation across Scaling Law}
Table~\ref{tab:main_results} reports results on 7B/13B-sized Llama2 and OPT models. In experiments with the Llama2 70B version, we observe that in a zero-shot setting, AdaInfer matches or slightly exceeds the dense model while reducing computational costs by 10\% to 50\%. However, in the few-shot setting, despite similar reductions in computation, AdaInfer's accuracy shows a 1\% to 25\% drop across different tasks compared to the dense model, \ie without early exit. This calls for more feature engineering for larger models, such as the 70B or even larger scales. Improving AdaInfer to adapt to these larger models is a direction for our future research. The results of all LLMs using different classifiers are summarized in Table~\ref{tab: all qa res} and Table~\ref{tab: all classification and rule res} in Appendix~\ref{all llms results}. The best results are highlighted for each task in the current setting.

\subsection{Generalization Study}

In Tables~\ref{tab:main_results}, we randomly select 6 training datasets from the entire pool of task training sets, which altogether contain 71 sub-datasets, to train the \modelname classifier. Furthermore, to assess the generalization performance of the statistical classifiers, we conduct the following tests.
\begin{itemize}
    \item \textbf{Intra-Task Generalization.} Evaluating the sentiment task using a classifier trained on the sentiment training dataset.
    \item \textbf{Inter-Task Generalization.} 
    Testing sentiment using a classifier trained on the knowledge question-answering task's dataset.
    \item \textbf{Inter-Model Generalization.} Assessing the sentiment task on Llama2-13B using a classifier trained on Llama2-7B.
\end{itemize}

The results are presented in Table~\ref{tab:classification-generalization}. The SVM classifier exhibits satisfactory intra-task and inter-task generalization capabilities, consistent with the results presented in the main results. However, for the CRF classifier, training in an intra-task manner leads to premature termination of the LLM at very shallow layers, resulting in subpar performance. This could be attributed to insufficient feature selection, causing the CRF to overfit noise or local features in the training data. Additionally, due to variations in the logits distribution characteristics among different models, the inter-model classifier's performance shows moderate accuracy. In conclusion, based on the results from Tables~\ref{tab:main_results}  and~\ref{tab:classification-generalization}, SVM is the choice of classifier for \modelname.

\begin{table}
\centering
\caption{Generalization performance of statistic classifier on sentiment task on Llama2-7B (32 layers), Inter-Model refers to Llama2-13B~(40 layers).}
\resizebox{\linewidth}{!}{
\begin{tabular}{ccrrrr}
\toprule
Classifier & Generalization & Acc& Layers & Variance & FLOPs\\
\midrule
SVM & \multirow{2}{*}{Intra-Task} & 94.90 &18.15&0.45&60.58 \\
CRF & &0.00 &0.00&0.00&0.00 \\
\midrule
SVM & \multirow{2}{*}{Inter-Task} & 95.50&19.20 &4.40&63.80 \\
CRF & &94.90 &20.20&4.55 &66.87 \\
\midrule
SVM & \multirow{2}{*}{Inter-Model} &90.70 &20.60&3.70&54.55 \\
CRF & &87.75 &19.20&2.75 &51.09\\
\bottomrule
\end{tabular}
}
\label{tab:classification-generalization}
\end{table}

\subsection{Factor Study}
In response to the features identified in Section~\ref{ssec:features}, we conduct cross-validation. Given that the classifiers in the main results utilized basic features (\ie ``gap'', ``top prob''), we explore the impact of features such as the cosine similarities between the current block and the previous block, which encompasses the attention values (attn), multi-layer perceptron (mlp), and hidden states. Results in Table~\ref{factor study} show that attention values have no discernible impact on the results, while other features like mlp and hidden states have an adverse effect. This result is consistent with the trend shown in Figure~\ref{fig:featureAnalysis}. It is our understanding that logits can measure whether the model's current forward progress is sufficient, while changes in other features may involve various factors.

\begin{table}
    \centering
    \caption{Comparative analysis of SVM performance with incremental feature addition in sentiment and MMLU/anatomy tasks.}
    \resizebox{\linewidth}{!}{
    \begin{tabular}{c|ccc}
        \toprule
        Feature & Sentiment  & MMLU  \\
        \midrule
        Base Features (gap, top prob)& 94.90&41.13 \\
        +attn &94.90 &41.13 \\
        +hidden state &67.53 &41.13 \\
        +mlp &67.88 & 41.93\\
        \bottomrule
    \end{tabular}}
    \label{factor study}
\end{table}

\section{Conclusion}
In this paper, we first provide an analysis to show that not all layers are necessary during inference for LLMs. Then, we present \modelname, a simple yet effective algorithm that dynamically determines the appropriate moment to cease inference based on the input instance. The decision is predicted by a low-cost statistical classifier using two easily obtainable features: the probability estimated for the most likely token and the gap between this probability and that of the next most likely token. While these two features may not capture all the evidence needed for early exit, the results are very promising for enhancing inference efficiency and adaptability, without modifying the model's parameters. Experiments on well-known LLMs (e.g., Llama2 series and OPT) show that \modelname can achieve an average pruning ratio of 17.8\%, and up to 43\% on sentiment tasks, with nearly no performance drop (<1\%). The computational savings can be more significant if an LLM is deployed to process mostly simple tasks and only occasionally difficult ones. Furthermore, \modelname is compatible with other model acceleration techniques, potentially offering further improvements in inference efficiency. We argue that \modelname establishes a new paradigm for efficient inference alongside existing effective methods.

\section*{Limitations}

In this paper, we make the first attempt to discover that the logits of each block are critical for early-exit classifiers in LLMs, incorporating this insight as a key design choice in \modelname. However, since \modelname relies on a single forward pass, it has not yet been extended to sequential generative tasks, offering significant avenues for future research. Further, the may exisit more effective features in addition to logits. 

\section*{Ethics Statement}
Our research aims to optimize large-scale model inference without modifying parameters, promising efficiency gains and reduced energy consumption. However, we must address potential misuse concerns, as enhanced inference capabilities may also enable malicious actors to exploit large neural language systems by injecting or amplifying logits as features, leading to undesirable behavior.

\section*{Acknowledgments}

This work is supported by the National Science and Technology Major Project~(2022ZD0116300) and the National Science Foundation of China~(No. 62106249).

\bibliography{acl}

\appendix

\begin{figure*}
  \centering

  \begin{subfigure}{0.4\linewidth}
    \centering
    \includegraphics[width=\linewidth]{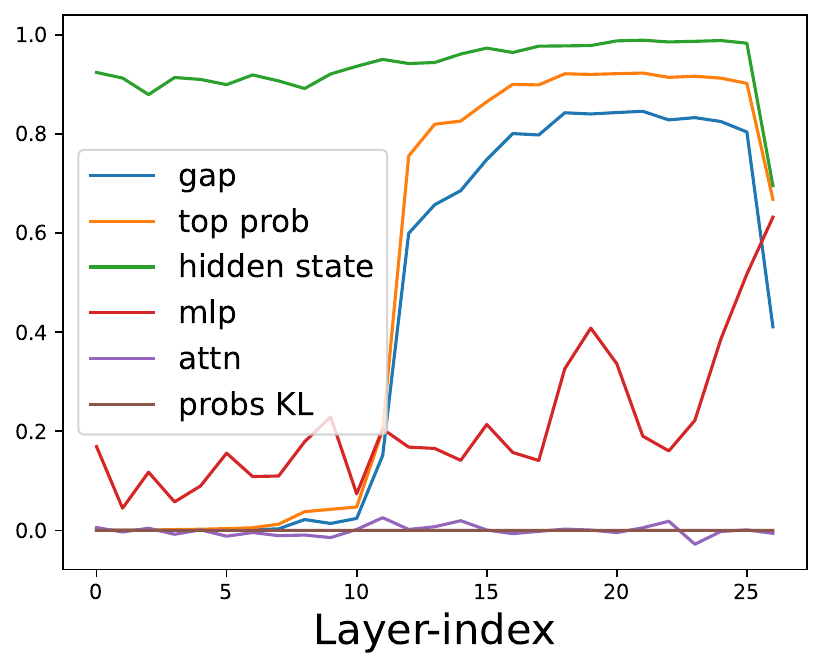}
    \caption{GPT-J 6B on sentiment}
  \end{subfigure}
  \begin{subfigure}{0.4\linewidth}
    \centering
    \includegraphics[width=\linewidth]{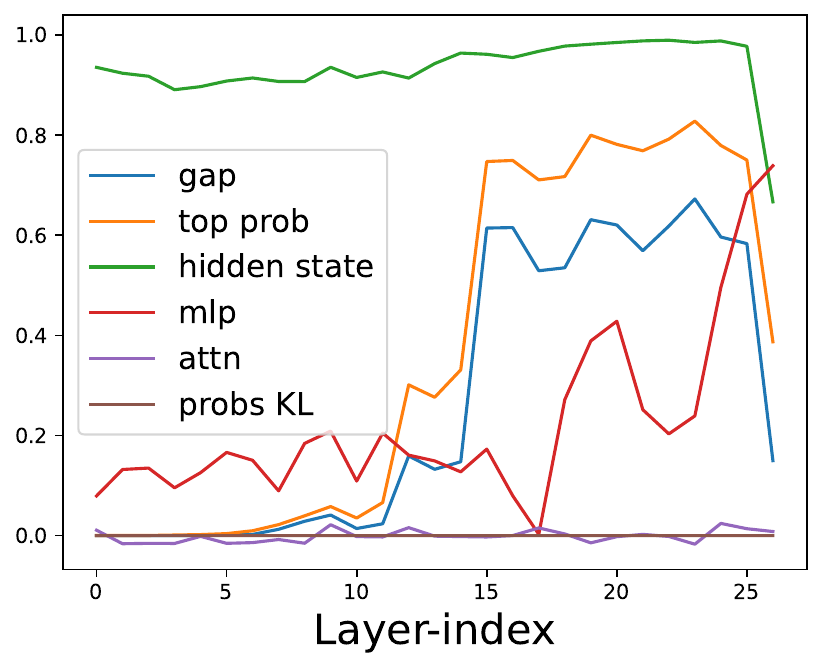}
    \caption{GPT-J 6B on MMLU}
  \end{subfigure}

  \begin{subfigure}{0.4\linewidth}
    \centering
    \includegraphics[width=\linewidth]{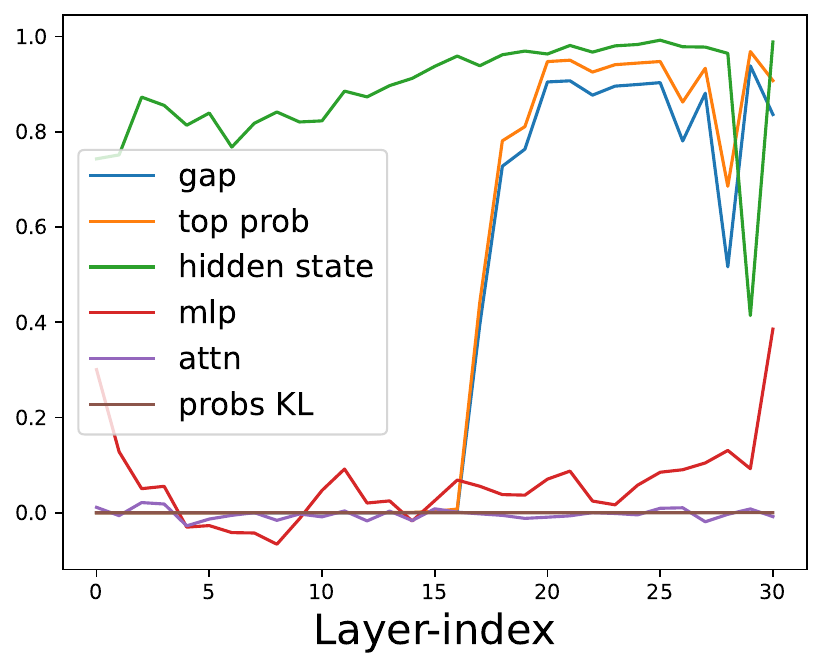}
    \caption{Llama2-7B on sentiment}
  \end{subfigure}
  \begin{subfigure}{0.4\linewidth}
    \centering
    \includegraphics[width=\linewidth]{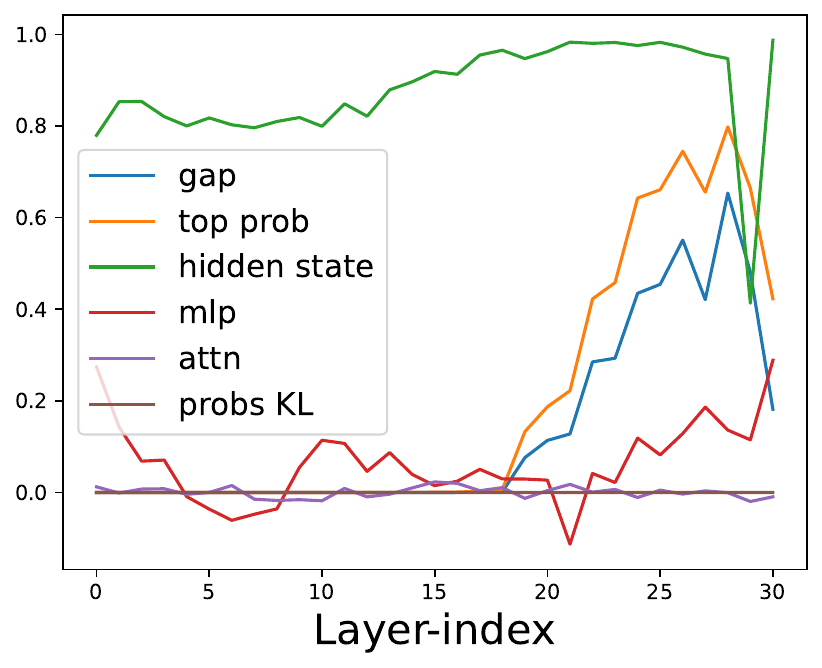}
    \caption{Llama2-7B on MMLU}
  \end{subfigure}

  \begin{subfigure}{0.4\linewidth}
    \centering
    \includegraphics[width=\linewidth]{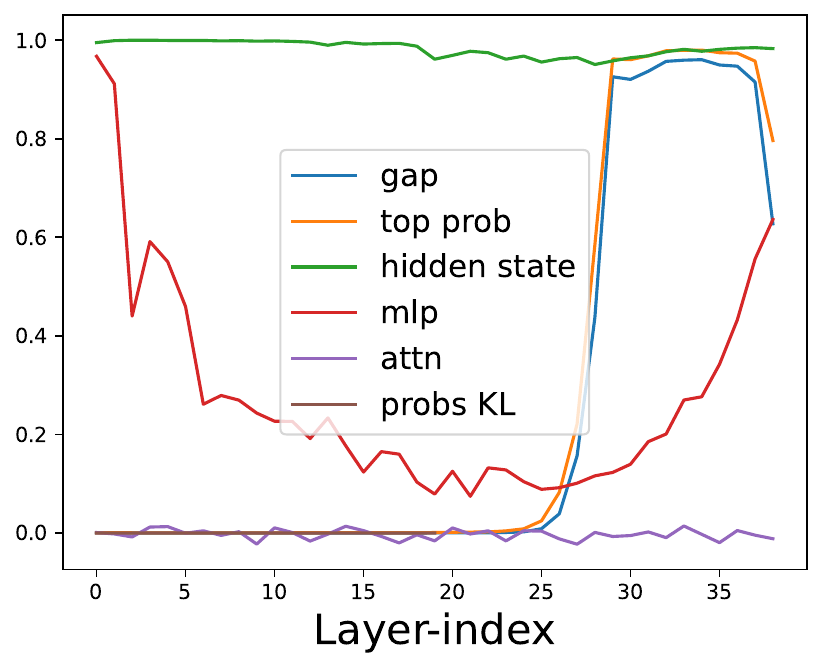}
    \caption{OPT-13B on sentiment}
  \end{subfigure}
  \begin{subfigure}{0.4\linewidth}
    \centering
    \includegraphics[width=\linewidth]{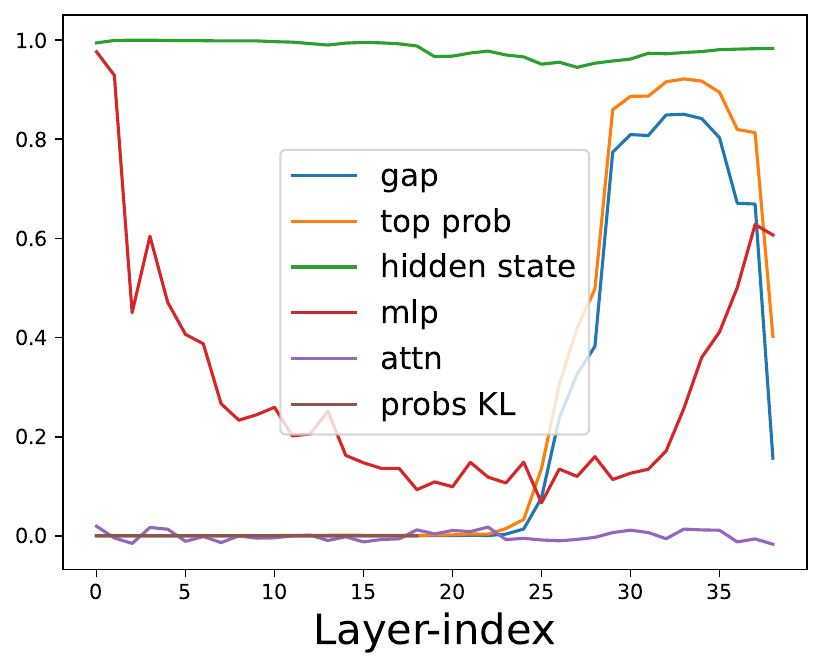}
    \caption{OPT-13B on MMLU}
  \end{subfigure}

  \caption{Visual analysis of diverse features across mainstream LLMs, on sentiment and MMLU tasks.}
  \label{fig: other llms}
\end{figure*}

\section{Computation Cost.} 

\paragraph{Classifier Computation Cost.} 
\label{svm cost}
We utilized the sklearn library for training SVM\footnote{\url{https://scikit-learn.org/stable/modules/svm.html}} and CRF\footnote{\url{https://sklearn-crfsuite.readthedocs.io/en/latest/}}, adhering to their default configurations.
Given a training dataset with $N$ training examples,
the time complexity for SVM training typically ranges from $O(N^2 \times d)$ to $O(N^3 \times d)$, where $d$ is the feature dimension. SVM prediction time complexity is $O(d)$ per single inference. For standard linear-chain CRF, the training time complexity is approximately $O(N \times S \times M)$, where $S$ is the average sequence length, $M$ is the label count. The prediction time complexity for CRF is $O(S \times M)$ per single inference.
In contrast, the inference time complexity for large models like llama2 is $LSd(d+S)$ per single inference, where $d$ is the hidden size, $S$ is the sequence length, and $L$ represents the number of layers. Comparatively, the computational load of SVM and CRF is negligible when compared to large models.

\paragraph{Transformer Computation Cost.}
\label{tf cost}
Given a language model with $l$ transformer layers, hidden size $h$, sequence length $s$, vocabulary size $V$, and batch size $B$, each transformer block needs $24Bsh^2+4Bs^2h$ FLOPs for a forward pass. 
The other main contributor to the FLOPs count is the classification layer in the language model head, which transforms features of dimension $h$ to the vocabulary dimension $V$. The required FLOPs for this operation is $2BshV$ in the forward pass. 
While \modelname does convert hidden states to logits at each block through classification layer, it only utilizes the hidden state from the last token for conversion, even when the sequence length is 2048 or longer. In the case of Llama2 7/13/70B, this computation accounts for only 0.000288, 0.000236, and 0.000152 of the total number of FLOPs for transformer inference. Similarly, for OPT 13B, it amounts to 0.000367. Consequently, the computational burden associated with this aspect can be disregarded.
Summing these together, a transformer model with $l$ transformer layers, the total number of floating-point operations for inference is $4Bshl(6h+s)+2BshV$. Thus, the ratio of inference cost in FLOPs can be calculated as
\begin{equation}
    \frac{2l'(6h+s)+V}{2l(6h+s)+V}
\end{equation}

\section{Classifier Training}
\label{sec:createTrainingData}

Considering a training input instance $x$ and its corresponding label $y$ from $D_{train}$. Once $x$ is processed through a decoder layer of LLM, we can extract a general feature vector $x^d$ ($d$ is the number of features). Additionally, we obtain the probability distribution $P$ over the vocabulary $V$ of the current layer's hidden state after passing through the classification layer (as depicted in Section~\ref{ssec:llmPrelim}). This can be represented as: $P=\softmax(WH+b)$, where $H$ is the hidden state of the current layer, $W$ and $b$ are the weights and bias of the classification layer, respectively. Function $\softmax$ is applied to convert logits to probabilities. Let the highest-ranked token in this distribution be $\hat{y} = \text{argmax}(P)$, where $\text{argmax}(P)$ finds the token with the highest probability. If $\hat{y}$ matches the label $y$, the associated label $y_c$ for the feature vector $x_d$ is designated as positive; otherwise, it is labeled as negative. 
Thus, for an $L-$layer LLM, each input instance $x$ yields $L$ pairs of $\langle x^d, y_c\rangle$.

\section{More Observation of LLMs} \label{more llm observation}

Figure \ref{fig: other llms} depicts a visual analysis of features across the layers within each block of mainstream LLMs. It shows that the ``gap'' and ``top prob'' exhibit a gradual increase during the inference phase, reaching stability in the deeper layers. Additionally, the activation of ``gap'' and ``top prob'' varies across layers for different tasks. These observed trends align with the findings discussed in Section~\ref{ssec:features}.

\section{Comprehensive Summary of Results} 
\label{all llms results}

The results of all LLMs using different classifiers are summarized in Tables~\ref{tab: all qa res} and~\ref{tab: all classification and rule res}.
We  highlight the best results for each task in the current setting.\footnote{We report the top-1 accuracy score on the test set following function vectors \cite{todd2023function} (HELM implementation).} The experimental results indicate that (\romannumeral1) early exits are feasible for different tasks, (\romannumeral2) the timing of early exits varies depending on the instance, and (\romannumeral3) in both zero-shot and few-shot settings, accuracy is comparable with dense models. 
It's worth noting that for individual tasks, \modelname even outperforms the dense model in zero-shot or few-shot accuracy. This suggests that in inference scenarios, deep layers may tend to over-represent some tasks, potentially impairing performance.

\begin{table*}
    \centering
    \caption{Performance and computational efficiency in question answering tasks. Few-shot learning with sample sizes of \{5, 10, 15,  20\} are everaged. \modelname uses SVM, and \modelname \textit{w.} Rule using GAP treshold set at 0.8.}
    \resizebox{0.85\textwidth}{!}{
    \begin{tabular}{c c ccc ccc ccc}
        \toprule
        \multirow{2}{*}{Setting} & \multirow{2}{*}{Model} & \multicolumn{2}{c}{MMLU} & \multicolumn{2}{c}{CommonsenseQA} & \multicolumn{2}{c}{SQuAD} & \multicolumn{2}{c}{Avg} \\
        \cmidrule(lr){3-4} \cmidrule(lr){5-6} \cmidrule(lr){7-8} \cmidrule(lr){9-10}
        && Acc$\uparrow$ & FLOPs$\downarrow$ & Acc$\uparrow$ & FLOPs$\downarrow$ & Acc$\uparrow$ & FLOPs$\downarrow$ & Acc$\uparrow$ & FLOPs$\downarrow$\\
        \midrule
        \multirow{4}{*}{Zero-shot}
        & OPT-13B &7.95&100&\cellcolor{gray!20}8.20&100&20.00&100 &\cellcolor{gray!20}12.05&100\\
        &\modelname \textit{w.} Rule &3.21&\cellcolor{gray!20}89.58 &0.60& \cellcolor{gray!20}85.17&20.72&\cellcolor{gray!20}87.98 &8.18&\cellcolor{gray!20}87.58\\
        &\modelname \textit{w.} CRF &7.14&96.57&4.60&93.26&\cellcolor{gray!20}24.36&93.22&12.03&94.35\\
        &\modelname
        &\cellcolor{gray!20}8.67&97.55&2.80&97.55&23.00&97.55&11.49&97.55\\
        \cmidrule{2-10}
        \multirow{4}{*}{Few-shot}
        & OPT-13B  &\cellcolor{myblue}23.60 &100 & 21.45&100 & \cellcolor{myblue}26.12 &100 &23.72&100\\
        &\modelname \textit{w.} Rule &20.99 &\cellcolor{myblue}79.54 & 20.72&\cellcolor{myblue}80.00 & 24.20&\cellcolor{myblue}82.93&21.97&\cellcolor{myblue}80.82 \\
      &\modelname \textit{w.} CRF &24.44 &97.43 &21.18 &97.55 & 25.98&97.11&\cellcolor{myblue}24.81&97.37\\
        &\modelname & 22.59 &83.94 &\cellcolor{myblue}21.62 &86.05 &25.95 &88.31 &23.39&86.10\\
        \midrule 
        \multirow{4}{*}{Zero-shot}
        & Llama2-7B &4.19&100&\cellcolor{gray!20}5.30&100&20.40&100&9.96&100 \\
        &\modelname \textit{w.} Rule &4.69&\cellcolor{gray!20}95.69&4.60&\cellcolor{gray!20}94.90&\cellcolor{gray!20}23.90&\cellcolor{gray!20}89.48&11.06&\cellcolor{gray!20}93.36\\
        &\modelname \textit{w.} CRF &\cellcolor{gray!20}4.86&95.32&2.00&95.01&18.80&91.17&8.55&93.83\\
        &\modelname&4.63&96.13&4.80&95.26&23.80&89.98& \cellcolor{gray!20}11.08&93.79\\
        \cmidrule{2-10}
        \multirow{4}{*}{Few-shot}
        &Llama-2-7B &43.05 &100 &\cellcolor{myblue}53.50 &100 & \cellcolor{myblue}48.08&100&\cellcolor{myblue}48.21&100 \\
        &\modelname \textit{w.} Rule & \cellcolor{myblue}44.03&\cellcolor{myblue}93.69 &52.83 &\cellcolor{myblue}90.23 &45.68 &\cellcolor{myblue}86.72&47.51&\cellcolor{myblue}90.21\\
        &\modelname \textit{w.} CRF &41.38 &94.23 &53.6 &91.61 &43.62 &88.10 &46.20&91.31\\
        &\modelname &43.73 &93.76 &53.00 &90.46&45.82 &87.06&47.52&90.43 \\
        \midrule
        \multirow{4}{*}{Zero-shot}
        & Llama2-13B &2.54&100&1.00&100&19.20&100&7.58&100 \\
        &\modelname \textit{w.} Rule
        &\cellcolor{gray!20}5.35&\cellcolor{gray!20}90.84&1.10&\cellcolor{gray!20}92.78&24.60&\cellcolor{gray!20}73.17&\cellcolor{gray!20}10.35&\cellcolor{gray!20}85.60\\
        &\modelname \textit{w.}CRF
        &4.77&97.40&\cellcolor{gray!20}1.40&97.28&23.10&93.03&9.76&95.90\\
        &\modelname
        &2.48&98.14&0.70&98.37&\cellcolor{gray!20}25.90&85.34&9.69&93.95\\
        \cmidrule{2-10}
        \multirow{4}{*}{Few-shot}
        &Llama-2-13B &\cellcolor{myblue}53.31 &100&\cellcolor{myblue}64.92 &100&\cellcolor{myblue}52.9 &100&\cellcolor{myblue}57.04&100\\
        &\modelname \textit{w.} Rule 
        &47.09 &\cellcolor{myblue}84.10 &55.33 & \cellcolor{myblue}79.57&43.43 &\cellcolor{myblue}71.19 &48.62&\cellcolor{myblue}78.29\\
        &\modelname \textit{w.}CRF & 52.72 &97.15 &65.72 &96.40 & 51.75&89.94 &56.73&94.50\\
        &\modelname &52.44 &93.55 &62.48 &89.10 &48.35 &80.66 &54.42&87.77\\
        \bottomrule
    \end{tabular}
    }
    \label{tab: all qa res}
\end{table*}

\begin{table*}
    \centering
    \caption{Performance and computational efficiency in text classification and rule understanding tasks, with the same settings as the question answering task.}
    \resizebox{0.85\textwidth}{!}{
    \begin{tabular}{c c ccc ccc ccc}
        \toprule
        \multirow{2}{*}{Setting} & \multirow{2}{*}{Model} & \multicolumn{2}{c}{Sentiment} & \multicolumn{2}{c}{AG News} & \multicolumn{2}{c}{Avg} & \multicolumn{2}{c}{Rule Understanding} \\
        \cmidrule(lr){3-4} \cmidrule(lr){5-6} \cmidrule(lr){7-8} \cmidrule(lr){9-10}
        && Acc$\uparrow$ & FLOPs$\downarrow$ & Acc $\uparrow$& FLOPs$\downarrow$ & Acc$\uparrow$ & FLOPs$\downarrow$ & Acc$\uparrow$ & FLOPs$\downarrow$\\
        \midrule
        \multirow{4}{*}{Zero-shot}
        & OPT-13B &0.00&100&0.10&100&0.05&100&3.38&100\\
        &\modelname \textit{w.} Rule &0.00&\cellcolor{gray!20}90.61&0.10&\cellcolor{gray!20}92.03&0.05&\cellcolor{gray!20}91.32&3.64&\cellcolor{gray!20}87.55\\
        &\modelname \textit{w.} CRF &0.00&97.55&0.10&97.55&0.05&97.55&\cellcolor{gray!20}4.11&97.55\\
        &\modelname &0.00&96.87&0.10&100&0.05&98.44&3.86&92.52\\
        \cmidrule{2-10}
        \multirow{4}{*}{Few-shot}        
        &OPT-13B  &92.58 &100 & \cellcolor{myblue}72.83&100 & 82.71&100& \cellcolor{myblue}58.48 &100\\
        &\modelname \textit{w.} Rule &\cellcolor{myblue}94.20 &\cellcolor{myblue}78.30 & 12.95&\cellcolor{myblue}82.54 &53.58&\cellcolor{myblue}80.42&48.20&\cellcolor{myblue}85.50\\
        &\modelname \textit{w.} CRF &92.88 &97.50 &71.27 & 97.55& 82.08&97.53&55.33 &97.50\\
        &\modelname &92.97 &80.28 &\cellcolor{myblue}72.83 &100 & \cellcolor{myblue}82.90&90.14 &52.83 &89.74\\
        \midrule
        \multirow{4}{*}{Zero-shot}
        & Llama2-7B &0.00&100&0.10&100&0.05&100&\cellcolor{gray!20}5.47&100\\
        &\modelname \textit{w.} Rule &0.00&96.08&0.10&\cellcolor{gray!20}91.05&0.05&\cellcolor{gray!20}93.57&5.41&\cellcolor{gray!20}91.20\\
        &\modelname \textit{w.} CRF &0.00&\cellcolor{gray!20}96.07&0.10&92.20&0.05&94.14&3.62&92.08\\
        &\modelname &0.00&96.37&0.10&91.36&0.05&93.87&5.32&91.55\\
        \cmidrule{2-10}
        \multirow{4}{*}{Few-shot}
        &Llama-2-7B &95.20 &100&79.65 &100&87.43&100&66.80&100 \\
        &\modelname \textit{w.} Rule & \cellcolor{myblue}95.30&\cellcolor{myblue}67.78 &\cellcolor{myblue}79.72 &\cellcolor{myblue}94.38 &\cellcolor{myblue}87.51&\cellcolor{myblue}81.08&66.80&\cellcolor{myblue}87.99\\
        &\modelname \textit{w.} CRF &94.90 &69.91 &61.62 &96.38 &78.26&83.15&62.36&89.60 \\
        &\modelname &\cellcolor{myblue}95.30 &68.05 &\cellcolor{myblue}79.72 &94.51&\cellcolor{myblue}87.51&81.28&\cellcolor{myblue}66.92&88.41\\
        \midrule
        \multirow{4}{*}{Zero-shot}
        & Llama2-13B &0.00&100&0.10&100&0.05&100&2.32&100\\
        &\modelname \textit{w.} Rule &0.00&\cellcolor{gray!20}88.25&0.10&\cellcolor{gray!20}77.82&0.05&\cellcolor{gray!20}83.04&\cellcolor{gray!20}9.9&\cellcolor{gray!20}74.80\\
        &\modelname \textit{w.} CRF &0.00&97.27&0.10&94.04&0.05&95.66&3.43&90.29\\
        &\modelname &0.00&97.43&0.10&88.37&0.05&92.90&6.14&85.76\\
        \cmidrule{2-10}
        \multirow{4}{*}{Few-shot}
        &Llama-2-13B &\cellcolor{myblue}95.90 &100& \cellcolor{myblue}77.53&100&\cellcolor{myblue}86.72&100&\cellcolor{myblue}69.36&100 \\
        &\modelname \textit{w.} Rule &91.45&\cellcolor{myblue}51.25&69.17&\cellcolor{myblue}70.65&80.31&\cellcolor{myblue}60.95&53.78&\cellcolor{myblue}70.38\\
        &\modelname \textit{w.} CRF
        &95.60 &73.07 &76.77&93.08 &86.19&83.08&65.82&90.29\\
        &\modelname
        &92.65&59.70&76.43&87.69&84.54&73.70&61.87&80.61\\
        \bottomrule
    \end{tabular}
    }
    \label{tab: all classification and rule res}
\end{table*}

\end{document}